\documentclass[runningheads]{llncs}
\usepackage{bbding}
\usepackage{pifont}
\usepackage{amssymb}
\usepackage{amsmath}
\usepackage{graphicx}
\usepackage{subcaption}
\usepackage{xcolor}
\usepackage{multirow}

\begin{document}
\title{Adversarial Magnification to Deceive Deepfake Detection through Super Resolution}
\titlerunning{Adversarial Magnification to Deceive Deepfake Detection through SR}
\author{Anonymous Submission}
%
%
\author{Davide Alessandro Coccomini\inst{1,2}\orcidID{0000-0002-0755-6154} \and
Roberto Caldelli\inst{3,4}\orcidID{0000-0003-3471-1196} \and
Giuseppe Amato\inst{1}\orcidID{0000-0003-0171-4315} \and
Fabrizio Falchi\inst{1}\orcidID{0000-0001-6258-5313} \and
Claudio Gennaro\inst{1}\orcidID{0000-0002-3715-149X}}

\authorrunning{D. Coccomini et al.}
%
\institute{ISTI-CNR, Pisa, Italy \and University of Pisa, Pisa, Italy \and
CNIT, Florence, Italy \and Mercatorum University, Rome, Italy \\
\email{davidealessandro.coccomini@isti.cnr.it, roberto.caldelli@unifi.it, giuseppe.amato@isti.cnr.it, fabrizio.falchi@isti.cnr.it, claudio.gennaro@isti.cnr.it}}
\maketitle              

\begin{abstract}
Deepfake technology is rapidly advancing, posing significant challenges to the detection of manipulated media content. Parallel to that, some adversarial attack techniques have been developed to fool the deepfake detectors and make deepfakes even more difficult to be detected. This paper explores the application of super resolution techniques as a possible adversarial attack in deepfake detection. Through our experiments, we demonstrate that minimal changes made by these methods in the visual appearance of images can have a profound impact on the performance of deepfake detection systems.
We propose a novel attack using super resolution as a quick, black-box and effective method to camouflage fake images and/or generate false alarms on pristine images. Our results indicate that the usage of super resolution can significantly impair the accuracy of deepfake detectors, thereby highlighting the vulnerability of such systems to adversarial attacks. The code to reproduce our experiments is available at: \url{https://github.com/davide-coccomini/Adversarial-Magnification-to-Deceive-Deepfake-Detection-through-Super-Resolution}

\end{abstract}
\section{Introduction}
Manipulating content to spread misinformation and damage the reputation of people has never been easier than nowadays. We are witnessing the unstoppable evolution of those known as Deepfakes. These are counterfeit media contents which often show people saying or doing things they never actually said or did, distorting reality. Distinguishing pristine contents from manipulated ones is extremely difficult. For this reason, various deepfake detectors have been developed. These are, however, subject to various issues such as the need to be up-to-date to keep up with the latest deepfake generation methods or the ability to handle real-world situations. It is precisely in real-world contexts that deepfake detection systems could be faced with targeted attacks made to deceive them. Known as adversarial attacks, these are techniques that introduce noise or adversarial patches, specifically crafted to deceive the detector. Although they can also be very effective, these techniques may require deep knowledge of the deepfake detector they are trying to fool. In this paper, we attempt to exploit a Super Resolution (SR) technique, to camouflage deepfake images in a quick and black-box manner (in the sense that the attack is model-agnostic). Our approach allows us to cause a significant increase in the False Negative Rate (fake samples classified as pristine) of up to $18\%$. We also have shown how the usage of SR on pristine images can cause a drastic increase in false alarms of up to $14\%$, highlighting the inadequacy of some deepfake detectors, which will probably arise as these techniques continue to proliferate.

\section{Related Works}

\subsection{Deepfake Generation and Detection}
The generation of deepfakes involves the use of techniques that manipulate human faces to achieve realistic alterations in appearance or identity. Two primary approaches are commonly employed: Variational AutoEncoders (VAEs) and Generative Adversarial Networks (GANs). VAE-based methods utilize encoder-decoder pairs to decompose and recompose distinct faces. On the other hand, GAN-based methods use a discriminator to distinguish real and fake images, paired with a generator that creates fake faces to fool the discriminator. Notable Deepfake generation methods include Face2Face\cite{face2face} and FaceSwap\cite{faceswap}.
As deepfakes becomes more credible, there is a growing demand for systems capable of detecting them. To address this problem various deepfake detectors have been developed. Some methods are capable of analyzing deepfake videos by considering also the temporal information\cite{mintime,Baxevanakis2022TheMD,ftcn,CALDELLI202131} but most approaches focus on frame-based classification, evaluating each video frame individually\cite{coccominicombining,dfdc_solution} and being available to manage also simply still deepfake images. Also, competitions such as \cite{dolhansky2020deepfake} and \cite{jimaging8100263} have been organized to stimulate the resolution of this task.
The problem of deepfakes has also been extended to the detection of synthetic images in general such in \cite{coccominidiffusion,Dogoulis_2023,amoroso2023parents} increasing the variety of fake contents.

\subsection{Adversarial Attacks}
Adversarial attacks, such as noise addition and adversarial patches, exploit vulnerabilities in deepfake detectors to deceive them. Adversarial noise introduces subtle perturbations, while adversarial patches overlap patterns to trigger misclassification. The authors of \cite{huang2020fakeretouch} propose a framework called FakeRetouch, which aims to reduce artifacts in deepfake images without sacrificing image quality. By adding noise and using deep image filtering, they achieve high fidelity to the original deepfake images reducing the accuracy of deepfake detectors. In \cite{Hou_2023_CVPR} the authors propose a statistical consistency attack (StatAttack) against deepfake detectors by minimizing the statistical differences between natural and deepfake images through the addition of statistical-sensitive degradations.

\subsection{Super Resolution}
Super Resolution (SR) is a technique which aims to reconstruct a high-resolution version of a low-resolution image by utilizing information from multiple input images\cite{arefin2020multi} or by using prior knowledge about the relationship between high-resolution and low-resolution image pairs\cite{8723565,ledig2017photo}.
One of the main SR techniques is the one proposed in \cite{Lim_2017_CVPR_Workshops} where an Enhanced Deep Super Resolution network (EDSR) is presented; it introduces some improvements to the ResNet architecture for SR previously proposed in \cite{ledig2017photo}. They remove batch normalization layers to increase flexibility and reduce memory usage. They also propose the use of residual scaling layers to stabilize the training procedure. The model constructed with these modifications, and pre-trained with a lower upscaling factor was able to achieve good results in terms of convergence speed and final performance.

\section{The proposed attack}

\begin{figure}[t]
\setlength{\abovecaptionskip}{0pt}
  \centering
    \caption{SR attack pipeline: pre-processing and attack phases. The face size is reduced by a factor $K$, then restored to its original resolution using a SR algorithm and pasted back onto the source frame.}
    \includegraphics[width=0.8\textwidth]{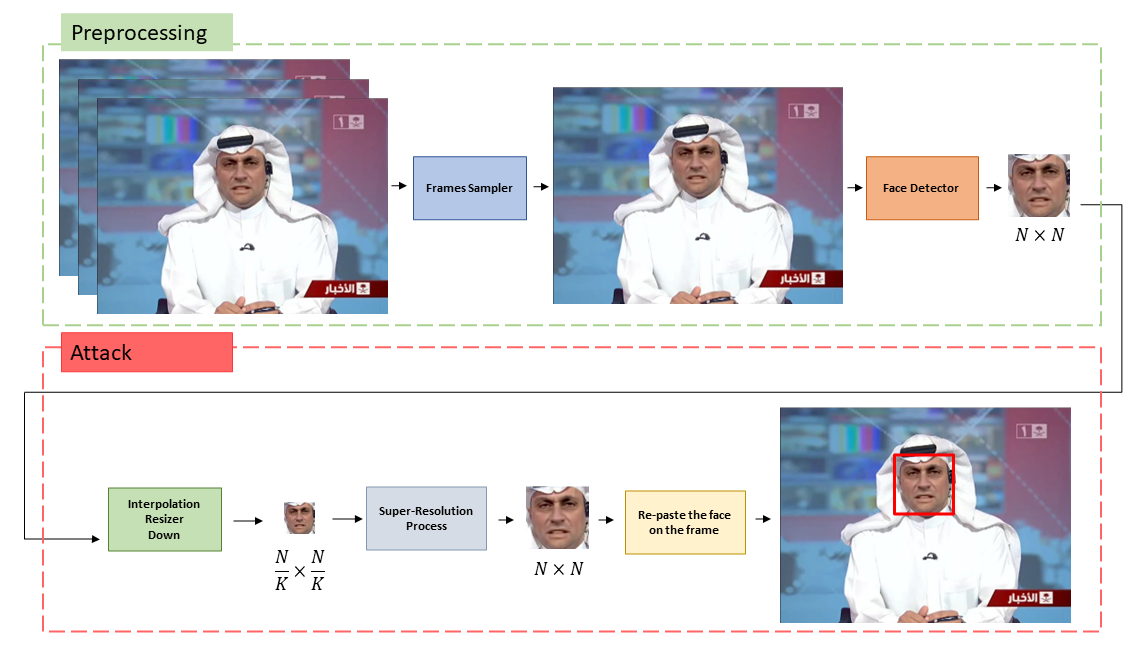}
  \label{fig:method}
  
\end{figure}
The proposed attack consists of exploiting SR techniques to modify a deepfake image and camouflage it in the eyes of a deepfake detector. The scope of the attack is then to mislead the deepfake detector and make the false negative rate increase. The SR process, in an attempt to improve the resolution of an image, could smooth the artifacts introduced by some deepfake generation techniques, thus undermining the learning performed by the deepfake detection model. Figure \ref{fig:method} shows the proposed framework for implementing the SR attack. Specifically, for each of the frames of a video (or for each image if the attack is applied to a still image) to be analyzed, a pretrained face detector (e.g., MTCNN\cite{mtcnn}) is applied. This step has been added to the pipeline for two main reasons. The first motivation is related to the scope of the attack itself since an attacker wants to manipulate a minimal part of the image in order to avoid adding artifacts when not needed. Applying the SR on the whole frame may add artifacts on the background finishing to have the inverse effect. The second reason behind the usage of a face detector is the common practice of both deepfake detectors and generators to focus only on the face and so it is very likely that the deepfake detector against which the attack is applied will only focus on the face and that the artifacts to be removed are concentrated on the face. 
The face extracted from the network has a specific resolution which is dependent on factors such as video resolution, distance from the camera, etc. 
Since the goal of SR is to raise the resolution of the image by a factor $K \in \mathbb{N}$, the image is firstly down-scaled by a factor $1/K$ and then given as input to an SR model (e.g. EDSR\cite{Lim_2017_CVPR_Workshops}) to be SR up-scaled by a factor K. The face image resulting from this process has the same size as the original detected one and so can be again put inside the source image from which it has been detected. So, to apply this method there is no need to know anything about the deepfake detector that will be used for the final detection, then the proposed method can be effectively considered a black-box attack and can be applied against any deepfake detector and on images manipulated with any deepfake generation method. Furthermore, this attack can also be carried out on deepfake content already generated and does not need to be integrated into the deepfake creation procedure.

\section{Experiments}

\subsection{Dataset}
Since we want to evaluate our attack on a variety of deepfake generation methods, we chose the well-known FaceForensics++ (FF++)\cite{rossler2019faceforensics} dataset for our experiments. The dataset consists of both pristine and manipulated videos created using various deepfake generation methods, namely Deepfakes\cite{deepfakes}, Face2Face\cite{face2face}, FaceShifter\cite{li2020advancing}, FaceSwap\cite{faceswap}, and NeuralTextures\cite{neuraltextures}. However, as this dataset consists of videos and the proposed attack exploits single-image SR, ten frames were randomly extracted for each of them on which face detection was then carried out. A training set and a test set were created for each deepfake generation method in FF++. Each training set consists of 14400 images, half of which were manipulated with one of the available methods. Each test set consists of 2800 images, half of which are manipulated again with the proposed attack. A total of five training and test sets are therefore available and all of them are perfectly balanced between the two classes (pristine and fake). To choose which videos should be used for training or test set we used the split made available in \cite{rossler2019faceforensics}.

\subsection{Experimental Setup}
To investigate the impact of the application of SR on the performance of deepfake detectors, we selected three architectures, namely Resnet50, Swin-Small and XceptionNet, and trained them on faces extracted from FF++ to perform a binary classification by pristine/fake image. For each training, the model only sees pristine images and fake ones manipulated with one of the available FF++ methods (SR is not applied). All models are pretrained on ImageNet and were fine-tuned with a learning rate of 0.01 for 30 epochs on an Nvidia Tesla T4. The test is carried out considering two different setups, in the first the models are tested by applying the SR-attack on both fake and pristine images. In the second, the pristine images are un-attacked and only the fake ones are passed through the SR process. The face is always extracted from each frame using a pretrained MTCNN\cite{mtcnn}. The scale factor used in our experiments is $K=2$ and so the extracted face is resized of a factor $1/K$ and then up-scaled through EDSR\cite{Lim_2017_CVPR_Workshops} restoring the original resolution. After this process, the face can be re-pasted to the frame exploiting the coordinates extracted during face detection.

\section{Results}

\subsection{Impact of Super Resolution on Deepfake Detection}

\begin{figure}[t]
\setlength{\abovecaptionskip}{5pt}
  \centering
  \caption{ROC curves on FF++ dataset for the three considered models.}
  \begin{subfigure}[b]{0.30\textwidth}
      \caption{Resnet50}
    \includegraphics[width=\textwidth]{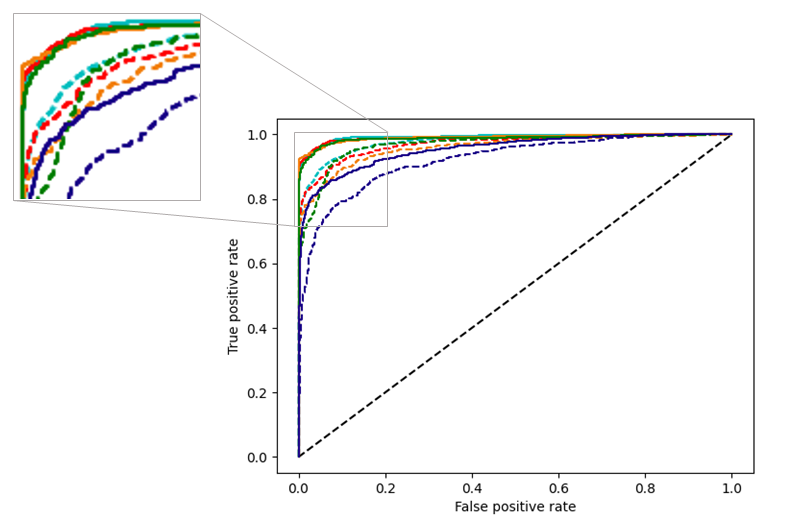}
    \label{fig:resnetroc}
  \end{subfigure}
  \begin{subfigure}[b]{0.30\textwidth}
  \caption{Swin}
    \includegraphics[width=\textwidth]{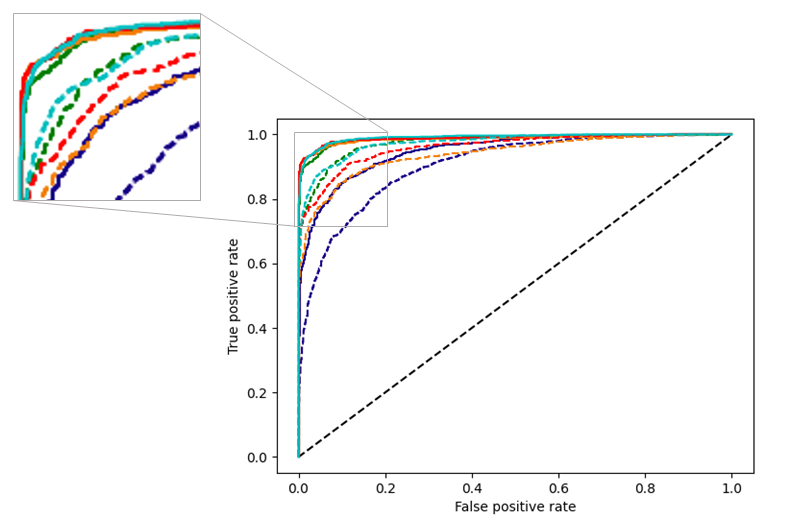}
    \label{fig:swinroc}
  \end{subfigure}
  \begin{subfigure}[b]{0.30\textwidth}
    \caption{XceptionNet}
    \includegraphics[width=\textwidth]{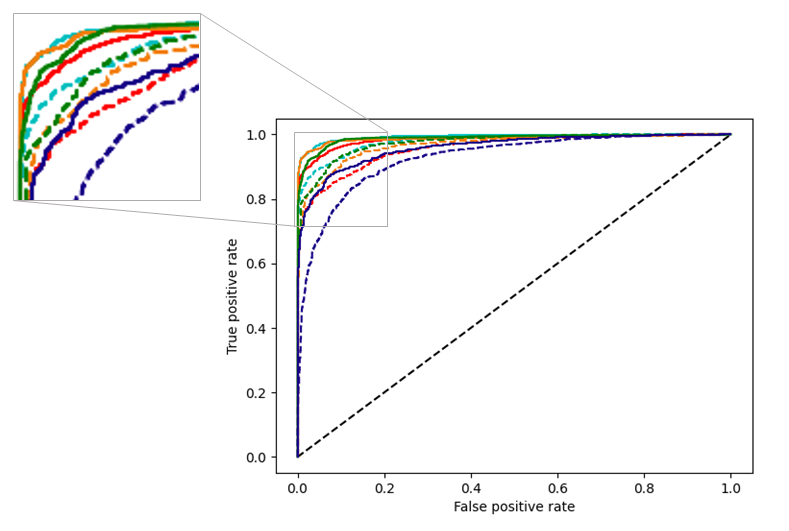}
    \label{fig:xceptionnetroc}
  \end{subfigure}
  
  \begin{subfigure}[b]{0.7\textwidth}
  \includegraphics[width=\textwidth]{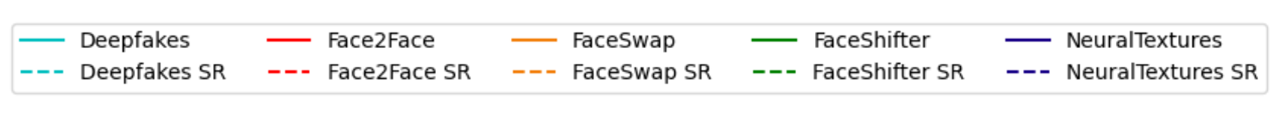}
  \end{subfigure}
  
  \label{fig:roc}
  \vspace{-9mm}
\end{figure}

\begin{table}[t]
    \centering
     \caption{Evaluation on FF++ test set (half pristine and half fake). The SR column indicates if the SR adversarial technique has been applied to the images. \textbf{Both pristine and fake images are attacked with SR.}}
    \setlength{\tabcolsep}{0.7em}
    \resizebox{\textwidth}{!}{%
    \begin{tabular}{l|l|rrrrrrr}

\hline\hline
    
    \textbf{Model} & \textbf{Forgery Method} & \textbf{SR} & \textbf{FNR} $\downarrow$ & \textbf{FPR} $\downarrow$ & \textbf{Recall} $\uparrow$& \textbf{Precision} $\uparrow$ & \textbf{AUC} $\uparrow$ & \textbf{Accuracy} $\uparrow$ \\
\hline
\multirow{10}{*}{\textbf{Resnet50}} & \multirow{2}{*}{Deepfakes} & $\times$ & 5.5 & 3.2 & 94.5 & 96.7 & 99.2 & 95.6 \\
& & $\checkmark$ & 6.9 & 10.1 & 93.1 & 90.2 & 97.7 & 91.5 \\
\cline{2-9}
& \multirow{2}{*}{Face2Face} & $\times$ & 5.0 & 3.2 & 95.0 & 96.7 & 98.9 & 95.9 \\
& & $\checkmark$ & 14.4 & 4.7 & 85.6 & 94.8 & 97.0 & 90.5 \\
\cline{2-9}
& \multirow{2}{*}{FaceSwap}  & $\times$ & 6.4 & 1.9 & 93.6 & 98.0 & 99.1 & 95.9 \\
& & $\checkmark$ & 21.1 & 2.6 & 78.9 & 96.8 & 96.0 & 88.1 \\
\cline{2-9}
& \multirow{2}{*}{FaceShifter} & $\times$ & 6.1 & 3.4 & 93.9 & 96.5 & 98.8 & 95.3 \\
& & $\checkmark$ & 24.8 & 3.3 & 75.2 & 95.8 & 96.8 & 86.0 \\
\cline{2-9}
& \multirow{2}{*}{NeuralTextures} & $\times$ & 14.1 & 8.1 & 85.9 & 91.3 & 95.4 & 88.9 \\
& & $\checkmark$ & 14.4 & 16.9 & 85.6 & 83.5 & 92.1 & 84.4 \\

\hline
\multirow{10}{*}{\textbf{Swin}} & \multirow{2}{*}{Deepfakes} & $\times$ & 5.9 & 3.6 & 94.1 & 96.3 & 99.1 & 95.3 \\
& & $\checkmark$ & 6.1 & 12.4 & 93.9 & 88.4 & 97.4 & 90.7 \\
\cline{2-9}
& \multirow{2}{*}{Face2Face} & $\times$ & 6.3 & 3.3 & 93.7 & 96.6 & 98.9 & 95.2 \\
&  & $\checkmark$ & 24.4 & 1.7 & 75.6 & 97.8 & 96.1 & 87.0 \\
\cline{2-9}
& \multirow{2}{*}{FaceSwap} & $\times$ & 4.9 & 4.6 & 95.1 & 95.3 & 98.6 & 95.2 \\
&  & $\checkmark$ & 21.9 & 5.3 & 78.1 & 93.7 & 93.9 & 86.4 \\
\cline{2-9}
& \multirow{2}{*}{FaceShifter} & $\times$ & 7.2 & 4.1 & 92.8 & 95.8 & 98.7 & 94.4 \\
&  & $\checkmark$ & 18.9 & 3.1 & 81.1 & 96.3 & 97.4 & 89.0 \\
\cline{2-9}
& \multirow{2}{*}{NeuralTextures} & $\times$ & 12.9 & 12.8 & 87.1 & 87.2 & 94.9 & 87.1 \\
&  & $\checkmark$ & 13.2 & 23.9 & 86.8 & 78.4 & 90.4 & 81.5 \\
\hline
\multirow{10}{*}{\textbf{XceptionNet}} & \multirow{2}{*}{Deepfakes} & $\times$ & 5.3 & 2.6 & 94.7 & 97.4 & 99.3 & 96.1 \\
& & $\checkmark$ & 5.6 & 12.4 & 94.4 & 88.4 & 97.9 & 91.0 \\
\cline{2-9}
& \multirow{2}{*}{Face2Face} & $\times$ & 9.6 & 3.3 & 90.4 & 96.5 & 98.4 & 93.6 \\
& & $\checkmark$ & 18.3 & 5.3 & 81.7 & 93.9 & 95.7 & 88.2 \\
\cline{2-9}
& \multirow{2}{*}{FaceSwap} & $\times$ & 5.1 & 3.2 & 94.9 & 96.7 & 98.8 & 95.8 \\
&  & $\checkmark$ & 15.8 & 4.9 & 84.2 & 94.5 & 96.6 & 89.6 \\
\cline{2-9}
& \multirow{2}{*}{FaceShifter} & $\times$ & 7.1 & 3.9 & 92.9 & 96.0 & 98.8 & 94.5 \\
&  & $\checkmark$ & 15.6 & 4.3 & 84.4 & 95.2 & 97.4 & 90.0 \\
\cline{2-9}
& \multirow{2}{*}{NeuralTextures} & $\times$ & 13.1 & 7.2 & 86.9 & 92.3 & 95.9 & 89.8 \\
& & $\checkmark$ & 9.8 & 21.6 & 90.2 & 80.7 & 92.7 & 84.3 \\
\hline
    \hline
    
    \end{tabular}
    }
\label{tab:exp1}
\end{table}
Table \ref{tab:exp1} shows the results obtained from the deep learning models considered to perform the deepfake detection task on the FF++ test set, with and without the usage of SR on both fake and pristine images. Observing the accuracy results on all the methods considered, the application of SR leads to a relevant drop in performance in all cases, confirming the hypothesis that the SR process can generate confusion in the deepfake detectors, thereby leading them to make errors. More in detail, looking at the False Negative Rate (FNR) and False Positive Rate (FPR) all the models seem to have a peak when the SR attack is applied.
When the deepfake generation method used on the image is Deepfakes or NeuralTextures, the impact on the FNR is less evident but the same detector that results in more robust on the fake images, fails on the pristine images attacked with SR and we see a huge increase in the FPR. The situation is exactly the opposite for the methods Face2Face, FaceSwap and FaceShifter on which the models seem to be more sensible on the fake images attacked with SR and so have an important increase on FNR while a slight swing in FPR is registered.
Increasing the FNR is the main interest for an attacker as it can be useful to be able to camouflage fake images against an automatic system that may be trying to filter them out. Vice versa, the increase in the FPR in some cases, highlights a serious problem in deepfake detection systems that, if SR became more widespread (e.g. on social media to improve the final visual quality), would end up confusing legitimate images for deepfakes and also open the door for an attacker to deliberately raise false alarms in the system.
That the use of SR pushes all Deepfake Detection models into error is also shown in Figure \ref{fig:roc} where it can be seen that in all cases, the AUCs obtained by the models on SR images are drastically lower (dashed lines) than their counterpart tested on images on which the SR process has not been applied (solid lines).

\begin{table}[t]
    \centering
    \caption{Evaluation on FF++ test set (half pristine and half fake). The SR column indicates if the SR adversarial technique has been applied to the images. \textbf{Only the fake images are attacked with SR.}}
   
    \setlength{\tabcolsep}{0.7em}
    \resizebox{\textwidth}{!}{%
    \begin{tabular}{l|l|rrrrrrr}
    \hline
    \textbf{Model} & \textbf{Forgery Method} & \textbf{SR} & \textbf{FNR} $\downarrow$& \textbf{FPR} $\downarrow$& \textbf{Recall} $\uparrow$ & \textbf{Precision} $\uparrow$ & \textbf{AUC} $\uparrow$ & \textbf{Accuracy} $\uparrow$\\
    \hline
    \hline
\multirow{10}{*}{\textbf{Resnet50}} & \multirow{2}{*}{Deepfakes} & $\times$ & 5.5 & \multirow{2}{*}{3.2} & 94.5 & 96.7 & 99.2 & 95.6\\
& & $\checkmark$ & 6.9 &  & 93.1 & 96.7 & 98.9 & 95.0\\
\cline{2-9}
& \multirow{2}{*}{Face2Face} & $\times$ & 5.0 & \multirow{2}{*}{3.2} & 95.0 & 96.7 & 98.9 & 95.9\\
&  & $\checkmark$ & 14.4 & & 85.6 & 96.4 & 97.6 & 91.2\\
\cline{2-9}
& \multirow{2}{*}{FaceSwap} & $\times$ & 6.4 & \multirow{2}{*}{1.9} & 93.6 & 98.0 & 99.1 & 95.9\\
& & $\checkmark$ & 21.1 &  & 78.9 & 97.6 & 95.6 & 88.5\\
\cline{2-9}
& \multirow{2}{*}{FaceShifter} & $\times$ & 6.1 & \multirow{2}{*}{3.4} & 93.9 & 96.5 & 98.8 & 95.3\\
& & $\checkmark$ & 24.8 &  & 75.2 & 95.7 & 95.2 & 85.9\\
\cline{2-9}
& \multirow{2}{*}{NeuralTextures} & $\times$ & 14.1 & \multirow{2}{*}{8.1} & 85.9 & 91.3 & 95.4 & 88.9\\
& & $\checkmark$ & 14.4 &  & 85.6 & 91.3 & 94.7 & 88.7\\
\hline
\multirow{10}{*}{\textbf{Swin}} & \multirow{2}{*}{Deepfakes} & $\times$ & 5.9 & \multirow{2}{*}{3.6} & 94.1 & 96.3 & 99.1 & 95.3\\
& & $\checkmark$ & 6.1 &  & 93.9 & 96.3 & 98.7 & 95.1\\
\cline{2-9}
& \multirow{2}{*}{Face2Face} & $\times$ & 6.3 & \multirow{2}{*}{3.3} & 93.7 & 96.6 & 98.9 & 95.2\\
& & $\checkmark$ & 24.4 &  & 75.6 & 95.8 & 96.2 & 86.2\\
\cline{2-9}
& \multirow{2}{*}{FaceSwap} & $\times$ & 4.9 & \multirow{2}{*}{4.6} & 95.1 & 95.3 & 98.6 & 95.2\\
& & $\checkmark$ & 21.9 &  & 78.1 & 94.4 & 94.1 & 86.7\\
\cline{2-9}
& \multirow{2}{*}{FaceShifter} & $\times$ & 7.2 & \multirow{2}{*}{4.1} & 92.8 & 95.8 & 98.7 & 94.4\\
& & $\checkmark$ & 18.9 &  & 81.1 & 95.2 & 96.6 & 88.5\\
\cline{2-9}
& \multirow{2}{*}{NeuralTextures} & $\times$ & 12.9 & \multirow{2}{*}{12.8} & 87.1 & 87.2 & 94.9 & 87.1\\
& & $\checkmark$ & 13.2 &  & 86.8 & 87.2 & 93.9 & 87.0\\
\hline
\multirow{10}{*}{\textbf{XceptionNet}} & \multirow{2}{*}{Deepfakes} & $\times$ & 5.3 & \multirow{2}{*}{2.6} & 94.7 & 97.4 & 99.3 & 96.1\\
& & $\checkmark$ & 5.6 &  & 94.4 & 97.3 & 99.2 & 95.9\\
\cline{2-9}
& \multirow{2}{*}{Face2Face} & $\times$ & 9.6 & \multirow{2}{*}{3.3} & 90.4 & 96.5 & 98.4 & 93.6\\
& & $\checkmark$ & 18.3 &  & 81.7 & 96.1 & 96.9 & 89.2\\
\cline{2-9}
& \multirow{2}{*}{FaceSwap} & $\times$ & 5.1 & \multirow{2}{*}{3.2} & 94.9 & 96.7 & 98.8 & 95.8\\
& & $\checkmark$ & 15.8 &  & 84.2 & 96.3 & 97.1 & 90.5\\
\cline{2-9}
& \multirow{2}{*}{FaceShifter} & $\times$ & 7.1 & \multirow{2}{*}{3.9} & 92.9 & 96.0 & 98.8 & 94.5\\
& & $\checkmark$ & 15.6 &  & 84.4 & 95.6 & 97.1 & 90.2\\
\cline{2-9}
& \multirow{2}{*}{NeuralTextures} & $\times$ & 13.1 & \multirow{2}{*}{7.2} & 86.9 & 92.3 & 95.9 & 89.8\\
& & $\checkmark$ & 9.8 &  & 90.2 & 92.6 & 96.5 & 91.5\\
    \hline
    \hline
    \end{tabular}
    }
\label{tab:exp2}
\end{table}
To evaluate deepfake detectors in a realistic context, an alternative test set was considered in which pristine images are not subjected to the SR process. In fact, an attacker has much more interest in generating false negatives than false positives, so as to go undetected by automated systems. As can be seen from the experiments reported in Table \ref{tab:exp2} in this setup the accuracy decreases, though more slightly, in almost all the cases with some deepfake generation methods on which the detectors are more robust to the attack. More in detail, the Face2Face, FaceSwap and FaceShifter images enhanced with the SR attack, are very difficult to detect, probably because the artifacts which the detector has learnt to recognize during the training process, are hidden by the SR process and this is translated in an higher FNR and a lower Recall value. In all the cases, the FPR is not affected by the usage of the SR attack since the pristine images are not attacked in this setup.  

\subsection{Visual Impact Analysis}
\label{sec:visual}
When performing an SR attack on a fake image, it is important that it remains as indistinguishable to human eyes as possible so as to preserve its meaning but also to make it less suspicious to users. To assess the impact of our attack on the image appearance, we compared the similarity of each image pair (non-SR, SR) through two commonly used quality metrics, \emph{Structural Similarity Index (SSIM)} and \emph{Peak Signal-to-Noise Ratio (PSNR)}. The SSIM is calculated as $\text{SSIM}(x, y) = \frac{{(2\mu_x\mu_y + C_1)(2\sigma_{xy} + C_2)}}{{(\mu_x^2 + \mu_y^2 + C_1)(\sigma_x^2 + \sigma_y^2 + C_2)}}$, where $x$ and $y$ are the two compared images, $\mu_x$ and $\mu_y$ are the average values of $x$ and $y$ respectively, $\sigma_x$ and $\sigma_y$ are the standard deviations, $\sigma_{xy}$ is the covariance between x and y and $C_1$ and $C_2$ are two constants used for stability. To calculate the PSNR we used the formula $\text{PSNR}(x, y) = 10 \cdot \log_{10}(\frac{{\text{MAX}^2}}{{\text{MSE}}(x, y)})$, where $x$ and $y$ are the two compared images, \text{MAX} is the maximum possible pixel value of the images and $\text{MSE}(x, y)$ is the Mean Squared Error between the images.

The values obtained from each image pair were used to calculate the mean to see the similarity between images attacked with and without SR for each category.
\begin{table}[t]
\centering
\caption{Evaluation of similarity between non-SR images and SR ones for each forgery method and for the pristine case.}
\vspace{1mm}
\label{tab:similarity}
\setlength{\tabcolsep}{4pt}
\begin{tabular}{l|c|c}
\hline
\hline
\textbf{Forgery Method} & \textbf{SSIM} $\uparrow$ & \textbf{PSNR $\uparrow$ (dB)} \\
\hline\hline
\emph{Pristine} & 0.968 & 39.8 \\
Deepfakes &  0.970 & 40.3 \\
Face2Face &  0.968 & 39.8 \\
FaceShifter & 0.973 & 40.9 \\
FaceSwap & 0.967 & 39.7\\
NeuralTextures & 0.972 & 40.5 \\
\hline
\hline
\end{tabular}
\end{table}
As can be seen from Table \ref{tab:similarity} the similarity between the SR images and the non-SR ones is very high, with SSIM values around  $0.97$ and PSNR around $40dB$ meaning a strong similarity and minimal changes brought by the SR process. We also checked if exists a correlation between the SSIM value and the variation in the error of the classifiers. In other words, we explored if a lower SSIM value is related to a higher number of misclassifications during the detection. From our experiments, in all the methods the correlation is lower than $\pm0.1$ meaning that the variation in detectors' performances is more related to the type of changes done to the image and not to the quantity of these. 

\subsection{Qualitative Evaluation}

To better understand the effect of the SR Attack on images, we visually analyzed some examples of deepfakes (e.g. Face2Face and FaceSwap) correctly detected by a Resnet50-based detector before the application of the attack but misclassified after it. 

\begin{figure}[htb!]
\setlength{\abovecaptionskip}{0pt}

  \centering
    \caption{Examples of fake images that are correctly detected by a Resnet50-based deepfake detector but not detected when the SR attack is applied.}
    \includegraphics[width=0.47\textwidth]{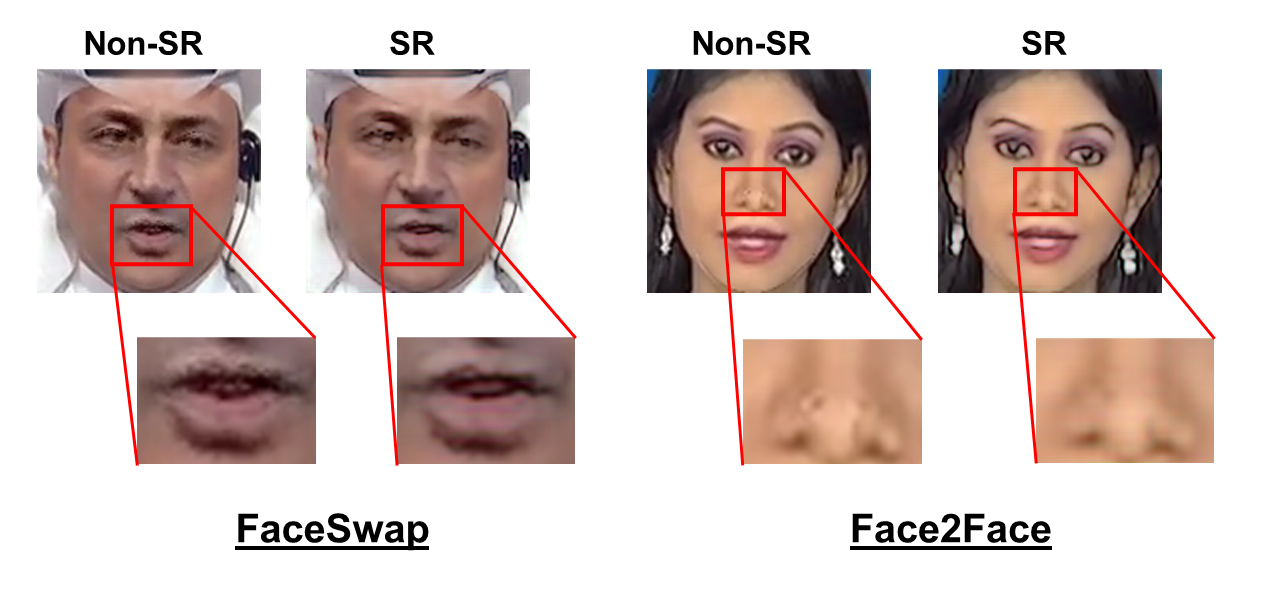}
  \label{fig:qualitative}
\end{figure}

These methods tend to introduce rather specific artifacts that, as visible in Figure \ref{fig:qualitative}, are then smoothed by the SR. This makes the work of the deepfake detector more difficult, as it has learnt to recognize such anomalies. As can be seen from the figure also the visual difference is minimal, as already stated by the analysis conducted in Section \ref{sec:visual}, but it is enough to make some artifacts around the mouth (FaceSwap) or on the nose (Face2Face) to disappear.

\section{Conclusions}
In this work, we examined the impact of applying SR on deepfake images in the context of deepfake detection. According to our experiments, the use of these techniques has a huge impact on the performance of deepfake detectors, causing the FNR to be drastically raised depending on the deepfake generation technique used and the artifacts introduced by it into the image. Also, a tendency was observed for deepfake detectors trained on specific deepfake generation methods to mistake pristine SR images for fake images when the SR attack is applied, causing the FPR to rise dramatically. In conclusion, SR attack can become an effective black-box attack in deepfake detection. In future work, we will explore the impact of detected face resolution on the attack performance, explore more SR techniques and also see if using SR as a data augmentation during the training process could be effective to make detectors robust to this attack.

\section*{Acknowledgments}
This work was partially supported by the project SERICS (PE00000014) under the NRRP MUR program funded by the EU - NGEU and by the H2020 project AI4Media (GA n. 951911).

%
%
%
\bibliographystyle{splncs04}
\bibliography{bibfile}

\end{document}